\documentclass[10pt,journal,compsoc]{IEEEtran}

\usepackage{subfig}
\usepackage{multirow}
\usepackage{pifont}
\usepackage{tikz}
\usepackage{tabularx}
\usepackage{hyperref}
\usepackage{amsfonts}   
\usepackage{mathtools}  
\usepackage{balance}    

\makeatletter
\newcommand{\thickhline}{%
    \noalign {\ifnum 0=`}\fi \hrule height 1pt
    \futurelet \reserved@a \@xhline
}
\newcommand{\myarrow}[1][]{%
  \begin{tikzpicture}[#1]%
    \draw (0,0.7ex) -- (0,0) -- (0.75em,0);
    \draw (0.55em,0.2em) -- (0.75em,0) -- (0.55em,-0.2em);
  \end{tikzpicture}%
}

\newcommand{\E}{\mathrm{E}}
\newcommand{\Var}{\mathrm{Var}}
\newcommand{\Cov}{\mathrm{Cov}}

\usepackage{xspace}
\newcommand{\dropout}{drop-conv2d\xspace}
\newcommand{\dropneuron}{drop-neuron\xspace}
\newcommand{\dropchannel}{drop-channel\xspace}
\newcommand{\droppath}{drop-path\xspace}
\newcommand{\droplayer}{drop-layer\xspace}

\newcommand{\Dropout}{Drop-Conv2d\xspace}
\newcommand{\Dropneuron}{Drop-Neuron\xspace}
\newcommand{\Dropchannel}{Drop-Channel\xspace}
\newcommand{\Droppath}{Drop-Path\xspace}
\newcommand{\Droplayer}{Drop-Layer\xspace}

\title{Efficient and Effective Dropout for Deep Convolutional Neural Networks}

\author{Shaofeng~Cai, Yao~Shu, Wei~Wang, Gang~Chen,~\IEEEmembership{Member,~IEEE},\\ Beng~Chin~Ooi,~\IEEEmembership{Fellow,~IEEE}, Meihui~Zhang~\IEEEmembership{Member,~IEEE}\\
\IEEEcompsocitemizethanks{
\IEEEcompsocthanksitem S. Cai, Y. Shu, W. Wang, B.C. Ooi are with
National University of Singapore, Singapore 117417.
E-mail: [shaofeng, shuyao95, ooibc, wangwei]@comp.nus.edu.sg
\IEEEcompsocthanksitem G. Chen is with Zhejiang University, Hangzhou 310027, China. E-mail: cg@cs.zju.edu.cn 
\IEEEcompsocthanksitem M. Zhang is with Beijing Institute of Technology, Beijing, China 100081. Email: meihui\_zhang@bit.edu.cn }
}

\begin{document}

\maketitle

\begin{abstract}

Convolutional Neural networks (CNNs) based applications have become ubiquitous, where proper regularization is greatly needed.
To prevent large neural network models from overfitting, dropout has been widely used as an efficient regularization technique in practice.
However, many recent works show that the standard dropout is ineffective or even detrimental to the training of CNNs.
In this paper, we revisit this issue and examine various dropout variants in an attempt to improve existing dropout-based regularization techniques for CNNs.
We attribute the failure of standard dropout to the conflict between the stochasticity of dropout and its following Batch Normalization (BN), and propose to reduce the conflict by placing dropout operations right before the convolutional operation instead of BN, or totally address this issue by replacing BN with Group Normalization (GN).
We further introduce a structurally more suited dropout variant \Dropout, which provides more efficient and effective regularization for deep CNNs.
These dropout variants can be readily integrated into the building blocks of CNNs and implemented in existing deep learning platforms.
Extensive experiments on benchmark datasets including CIFAR, SVHN and ImageNet are conducted to compare the existing building blocks and the proposed ones with dropout training.
Results show that our building blocks improve over state-of-the-art CNNs significantly, which can be ascribed to the better regularization and implicit model ensemble effect.



\end{abstract}


\section{Introduction}

Deep neural networks (DNNs) have achieved remarkable success in a variety of fields including computer vision~\cite{krizhevsky2012imagenet,he2016deep,hu2017squeeze}, natural language processing and healthcare analytics~\cite{dai2018fine,tracer}.
To increase model capacity for better performance, DNNs become larger and deeper with typically hundreds of layers and millions of parameters~\cite{he2016deep,huang2016densely,xie2017aggregated}.
However, large DNN models are prone to overfitting, and therefore proper regularization is greatly needed for the training of DNNs.

To improve generalization performance, many explicit and implicit regularization techniques are proposed, such as early stopping, weight decay, data augmentation~\cite{cutout} and etc.
Dropout~\cite{hinton2012improving,srivastava2014dropout} is arguably the most prominent regularization technique used in practice, due to its efficiency and effectiveness.
Specifically, for each training iteration, the standard dropout randomly samples a set of neurons and deactivates them, and then the training is conducted on the resultant subnet, which incurs negligible computational overhead.
Thereby, a new subnet is sampled and trained for each iteration.
The full network can thus be considered as an ensemble of an exponentially large number of subnets whose parameters are shared.
Besides the model ensemble effect, dropout also regularizes the networks by discouraging co-adaptation~\cite{srivastava2014dropout} between neurons and therefore contributes to more robust feature extraction.

However, recent attempts to apply dropout to convolutional neural networks~\cite{he2016deep,zagoruyko2016wide,huang2016densely} (CNNs) fail to obtain noticeable performance improvement.
Initially, dropout~\cite{srivastava2014dropout,hinton2012improving} is introduced to fully connected layers~\cite{krizhevsky2012imagenet}, which are however replaced by a global average pooling layer~\cite{lin2013network} thereafter.
Many attempts have also been made to apply dropout to convolution layers.
For instance, WRN~\cite{zagoruyko2016wide} applies a dropout layer between two wide convolution layers of the residual block and reports improved accuracy.
However, dropout for these CNNs is still adopted at the neuron level, which turns out to be less effective.
Even detrimental effects are observed~\cite{he2016identity} when introducing dropout to the identity mapping of the residual block in ResNet~\cite{he2016deep}.
The effectiveness of dropout for CNNs is further reduced by the introduction of other regularization techniques such as data augmentation and batch normalization~\cite{ioffe2015batch} in particular.

At a high level, the effectiveness of dropout training can be largely attributed to the regularization and ensemble effect engendered by random subnet sampling for training.
Following the same methodology, various dropout variants are also proposed injecting randomness into different CNN structural components, such as input patches~\cite{cutout}, connections~\cite{wan2013regularization}, neurons~\cite{srivastava2014dropout}, activation maps~\cite{tompson2015efficient,dropblock}, transformation paths~\cite{larsson2016fractalnet,xie2017aggregated}, residual blocks~\cite{huang2016deep,shakeshake}.
Although many of these variants can improve over the standard dropout to some extent, a closer and comprehensive investigation is much needed to provide  consistently efficient and effective dropout training for CNNs.

To better integrate dropout into CNNs, we revisit the existing dropout variants applied to different structural components, notably neurons, channels and paths.
In this paper, we present a unified framework to formulate and examine the three primary dropout variants, which are denoted as \Dropneuron, \Dropchannel and \Droppath respectively.
To uncover the reason behind the ineffectiveness of existing dropout methods for convolutional layers, we investigate the interaction between dropout and other common techniques adopted in CNNs, i.e., data augmentation and batch normalization~\cite{ioffe2015batch} (BN) in particular.
We find that \dropchannel is generally more effective in improving CNN training than \dropneuron.
We also note that for the neuron and channel level dropout, namely \dropneuron~\cite{li2019understanding} and \dropchannel, their conventional usage is in conflict with BN, which is adopted widely in CNNs to stabilize the first two moments (i.e., mean and variance) of its output distribution, whereas the random deactivation of the basic components (i.e., neurons and channels respectively)  with the \dropneuron and \dropchannel training disrupts such stability.
We then propose improved convolutional building blocks with \dropneuron/\dropchannel and common CNN layers, including dropout, convolution, batch normalization layer for ease of use (see Figure~\ref{fig:conv_formulation}).
We further examine various \droppath variants, i.e., dropout applied at the path level where transformation branches are randomly dropped during training~\cite{larsson2016fractalnet,xie2017aggregated,huang2016deep,shakeshake}, and find that path-dropping is largely effective in improving generalization.
Based on these observations, we integrate \dropchannel and \droppath and propose a more efficient and effective dropout variant \Dropout.
For convolutional layers trained with \dropout, each transformation branch between input and output channels is replicated $P$ times for larger model capacity, and these branches are randomly dropped during training to harness the dropout training benefits.
Then during inference, these branches can be readily merged back into one branch with no additional cost.

\begin{figure}[t]
    \centering
    \includegraphics[width=0.48\textwidth]{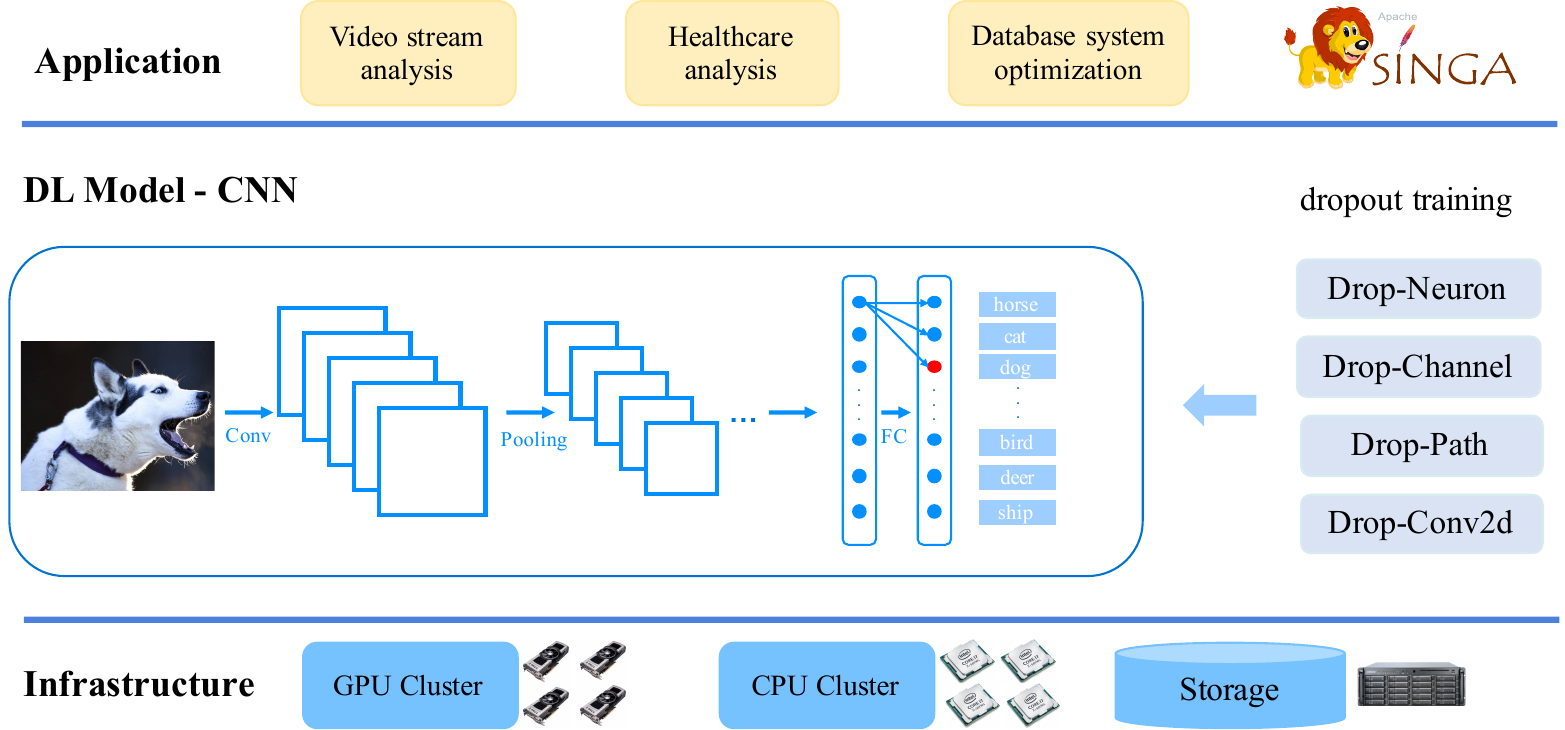}
    \caption{Supporting complex analytics with dropout training.}
    \label{fig:system}
\end{figure}

The advantages of the proposed building blocks are threefold.
Firstly, these dropout variants can be readily implemented as CNN building blocks in existing deep learning platforms such as Pytorch, TensorFlow or Apache SINGA\footnote{https://singa.apache.org/} initiated by us to support various DL-based applications, as illustrated in Figure~\ref{fig:system}.
Second, all these dropout variants are computationally lightweight and incur negligible computational overhead.
Third, the adoption of different dropout variants in convolutional layers of CNNs, especially \dropout, provides a more effective regularization.
To support these claims, extensive experiments are conducted over state-of-the-art CNNs.
We adopt widely benchmarked datasets CIFAR-10, CIFAR-100, SVHN and ImageNet, where significant accuracy improvement is observed even upon batch normalization and extensive data augmentation.
The main contributions can be summarized as follows:

\begin{itemize}
	\item We present a unified framework for analyzing dropout variants in CNNs. Specifically, we investigate the failure of \dropneuron and \dropchannel, which is mainly due to their conflict with BN in the convolutional block.
	\item We propose convolutional building blocks and \dropout, which are better in line with the dropout training mechanism and are readily applicable to existing CNN architectures.
	\item We conduct extensive experiments to examine different dropout variants and confirm the effectiveness of our proposed building blocks, and consistently achieve significant improvement for CNNs.
\end{itemize}

The remainder of the paper is organized as follows.
Section~\ref{sec:background} introduces the background.
In Section~\ref{sec:formulation}, we formulate the convolutional transformation in a unified framework, based on which training mechanisms of dropouts are examined, and improved building blocks and \dropout are proposed for deep CNNs.
Experimental evaluations of our proposed building blocks are provided in Section~\ref{sec:experiment}.
We conclude concludes this paper in Section~\ref{sec:conclusion}.

\section{Background}
\label{sec:background}

\subsection{Deep Convolutional Neural Networks}

One notable trend of recent convolutional neural networks (CNNs) is their growing depth and width.
AlexNet~\cite{krizhevsky2012imagenet} develops an 8 layer CNN for the large image classification dataset ImageNet and achieves great success.
The subsequent VGG~\cite{simonyan2014very} and GoogLeNet~\cite{szegedy2015going} push the depth of CNNs to 19 and 22 respectively by stacking their respective basic convolutional building blocks, e.g., Inception module in GoogLeNet.
ResNet~\cite{he2016deep} further proposes the residual connection that enables the training of ever deep CNNs over 1000 layers.
However, pushing CNN architectures deeper alone does not bring further benefits~\cite{he2016deep,veit2016residual}, and many CNN architectures instead grow wider~\cite{zagoruyko2016wide}, for larger model capacity and the ease of training.
\cite{lin2013network} replaces the filter kernel of the convolution with a multilayer perceptron, which facilitates interactions between input channels.
CNNs such as Inception series~\cite{szegedy2015going} and ResNeXt~\cite{xie2017aggregated} instead utilize group convolution~\cite{krizhevsky2012imagenet}, proposing the multi-branch convolution.
Other CNN building blocks of different connection topology and transformations are also proposed, such as Inception modules~\cite{szegedy2015going}, residual blocks~\cite{he2016deep,he2016identity}, dense connection blocks~\cite{huang2016densely} and remarkably various Neural Architecture Search blocks~\cite{nas,darts,understandnas} in recent years.

\subsection{Dropout for Deep Neural Networks}

Regularization is essential for deep neural networks, and many regularization methods have been proposed, such as weight decay, data augmentation, batch normalization~\cite{ioffe2015batch} and etc.
Dropout~\cite{srivastava2014dropout,hinton2012improving} is arguably the most prominent regularization technique used in practice.
\cite{wager2013dropout} shows that dropout performs a form of adaptive regularization for generalized linear models, which is first-order equivalent to an $L_2$ regularizer.
Further, dropout provides immediately the magnitude of the regularization, which is adaptively scaled by the inputs and the variance of the dropout variables~\cite{baldi2013understanding}.
\cite{gal2016dropout,kingma2015variational} analyze dropout in the Bayesian inference framework for uncertainty modeling.

Many relevant implicit model ensemble techniques based on dropout are also proposed.
Swapout generalizes dropout with a stochastic training method, which samples subnet for training from either dropout, stochastic depth~\cite{huang2016deep} or the residual connection~\cite{he2016deep}. 
DropConnect~\cite{wan2013regularization} instead introduces randomness to connections and randomly deactivates connections during training.
Model Slicing~\cite{modelslicing} randomly trains sliced subnets during training so that runtime accuracy-efficiency trade-offs can be achieved with these subnets.
More dropout variants are also proposed injecting randomness into different structural components.
SpatialDropout~\cite{tompson2015efficient} shows that adding one additional layer with dropout applied to channels can improve performance.
FractalNet~\cite{larsson2016fractalnet} proposes to randomly drops individual paths during training, and Stochastic depth~\cite{huang2016deep} randomly drops a subset of layers and forwards inputs with residual connection during training.
These dropout variants apply dropout to the basic components of CNNs, e.g., channels and paths, which regularizes CNNs for improved training.

\section{Dropout for Deep Convolutional Neural Networks}
\label{sec:formulation}

In this section, we first formulate the basic transformations of Convolutional Neural Networks from the viewpoint of split-transform-aggregate.
We then introduce general training mechanisms with dropout operations at different structural levels for CNNs.
We also examine various CNN architectures and propose convolutional building blocks with dropout training that are better in line with the dropout mechanisms for more efficient and effective regularization.

\subsection{The Basic Transformations of CNNs}

Broadly speaking, the topology of neural networks, including multi-layer perceptron, recurrent neural networks~\cite{hochreiter1997long} and convolutional neural networks~\cite{krizhevsky2012imagenet,he2016deep,huang2016densely}, can be represented precisely by a set of neurons and their connections from the connectionist viewpoint, where the information flow from input neurons to output neurons is regulated by learnable weights of each connection.
Succinctly, each neuron aggregates information from its input neurons by:

\begin{equation}
\label{formular:neuron_op}
y_i = \sum_{j=1}^{N} w_{ij} x_j
\end{equation}

where $\mathbf{x} = [x_1, x_2, \ldots, x_N]$ is a N-dimension input vector, and $w_{ij}$ the weight of the connection from input neuron $x_j$ to the output neuron $y_i$.
We omit bias and output nonlinearity here for brevity.
The neuron transformation follows the strategy of split-transform-aggregate, which can be interpreted as extracting features from all the input branches by first the dot product transformation of the input information with corresponding weights and then an aggregation over the input dimensions.

The transformation of convolutional neural networks can be formulated at a higher structural level with channels as the basic components instead of neurons.
Specifically, the most fundamental operation in CNNs comes from the convolutional layer which can be constructed to represent any given transformation $\mathcal{F}_{conv}:\mathbf{X} \rightarrow \mathbf{Y}$, where $\mathbf{X} \in \mathbb{R}^{C_{in} \times W_{in}\times H_{in}}$ is the input with $C_{in}$ channels of size $W_{in}\times H_{in}$, and $\mathbf{Y} \in \mathbb{R}^{C_{out} \times W_{out}\times H_{out}}$ is the output likewise. 

Denoting $\mathbf{X}, \mathbf{Y}$ as $[\mathbf{x}_1, \mathbf{x}_2, \ldots, \mathbf{x}_{C_{in}}]$, $[\mathbf{y}_1, \mathbf{y}_2, \ldots, \mathbf{y}_{C_{out}}]$ in vector of channels respectively, the parameter set associated with this convolutional layer comprises a series of filter kernels $\mathbf{W} = [\mathbf{w}_1, \mathbf{w}_2, \ldots, \mathbf{w}_{C_{out}}]$.
Then the convolutional transformation on $\mathbf{X}$ can be succinctly represented as:

\begin{equation}
\label{formular:conv_op}
\mathbf{y}_i = \mathbf{w}_i \ast \mathbf{X} = \sum_{j=1}^{C_{in}} \mathbf{w}_{i}^{j} \ast \mathbf{x}_j
\end{equation}

where $\ast$ denotes convolution operation, and $\mathbf{w}_i^j$ is a 2D spatial kernel associated with $i_{th}$ output channel $\mathbf{y}_i$ and convolves on $j_{th}$ input channel $\mathbf{x}_j$.
For a typical convolutional layer, each output channel $\mathbf{y}_i$ is connected to all the input channels $\mathbf{X}$ and is typically followed by some output nonlinearity, which is omitted here for succinctness.

\begin{figure*}[ht]
    \centering
    \includegraphics[width=0.7\textwidth]{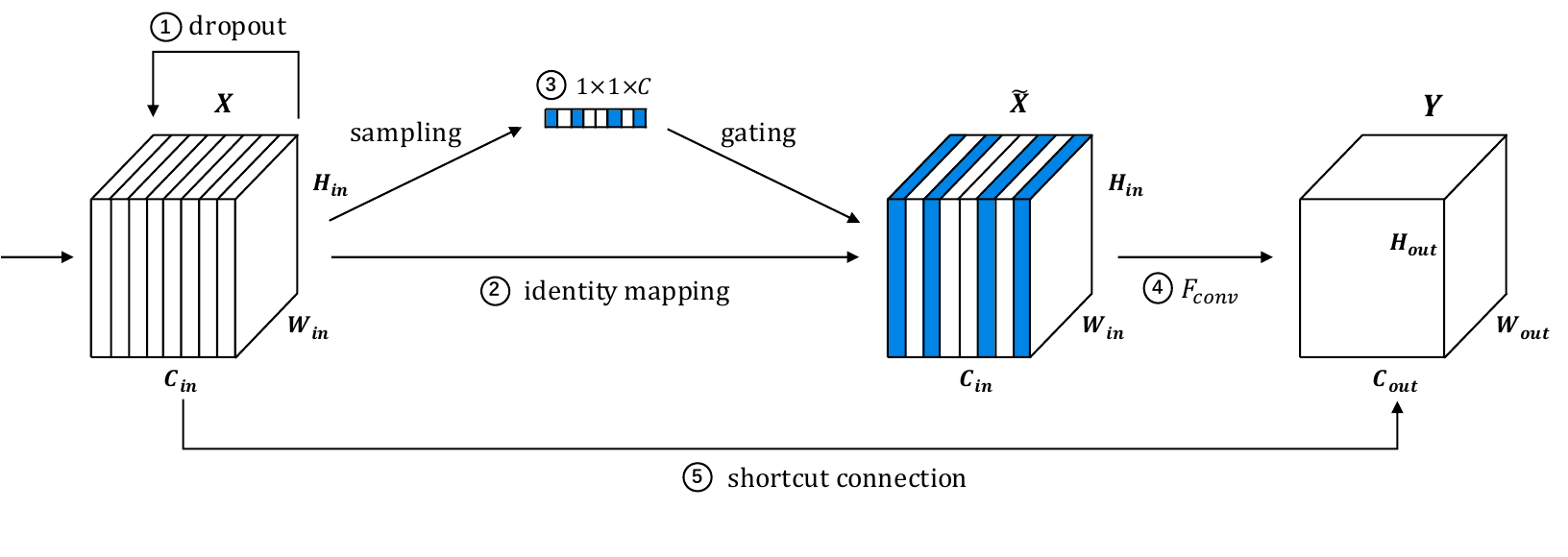}
    \caption{Illustration of various convolutional transformations. Dropout, or drop-neuron, gates input neurons in operation 1; Drop-channel replaces identity mapping in operation 2 with operation 3, random sampling and gating on channels; Drop-path is introduced to $\mathcal{F}_{conv}$ in operation 4 or to the shortcut connection in operation 5 (Drop-Layer).}
    \label{fig:conv_architecture}
\end{figure*}

We argue that the channel level representation more natural to the convolutional transformation of CNNs.
Structurally, CNNs consist of a stack of convolutional layers and for each convolutional layer, the transformation convolves over channels as in Equation~\ref{formular:conv_op}.
Unlike dense layers such as fully connected layers where each connection between input and output neurons is coupled with one learnable weight, neurons within the same channel share the same filter kernel for each output channel in CNNs.
Such a weight sharing strategy dramatically reduces the number of parameters of CNNs and suggests that CNNs extract features at the channel level.
This is also supported by the visualization of convolutional kernels~\cite{krizhevsky2012imagenet,zeiler2014visualizing}, where filter kernels of the input-connected layer learn to identify orientations and colored blobs with increasing invariance, and class discrimination is observed ascending the layers.

\subsection{Dropout Operations for CNNs}
\label{dropout_for_cnn}

In this subsection, we first outline mainstream convolutional transformations of representative CNN architectures.
Next, different structural levels of dropout operations are introduced and examined to improve the training of CNNs.
We then propose general convolutional building blocks integrated with built-in dropout operations and a more efficient and effective dropout variant \dropout for CNNs.

\subsubsection{Convolution Transformation Blocks}

We illustrate mainstream convolutional transformation in Figure~\ref{fig:conv_architecture}.
Conventional convolution transforms via the operation 2 and 4, namely an identity mapping of $\mathbf{X}$ and then a convolutional transformation $\mathcal{F}_{conv}$, which follows the formulation in Equation~\ref{formular:conv_op}.
Other convolution transformations such as group convolution~\cite{krizhevsky2012imagenet} and depth-wise convolution~\cite{howard2017mobilenets} can also be formulated accordingly with their respective constraints on channel connections.

Many convolutional transformations try to lengthen and/or widen the transformation.
For instance, NIN~\cite{lin2013network} lengthens $\mathcal{F}_{conv}$ by following the filter kernel with two layers of multilayer perceptron transformation, which is structurally equivalent to two convolutional layers with $1 \times 1$ filter.
Inception series~\cite{szegedy2015going,szegedy2017inception} widen $\mathcal{F}_{conv}$ with multiple heterogeneous transformation branches.
ResNeXt~\cite{xie2017aggregated} follows similar strategy by duplicating it $P$ times $\mathcal{F}_{conv}(\mathbf{X}) = \sum_{i=1}^{P} \mathcal{F}_{conv_\_i}(\mathbf{X})$, where the transformations are heterogeneous.
Other convolutional transformations encourage feature reuse by forwarding input channels $\mathbf{X}$ directly to output channels $\mathbf{Y}$, as is indicated in operation 5.
One commonly-adopted feature reuse is the shortcut of identity mapping proposed in ResNet~\cite{he2016deep,he2016identity}, i.e. $\mathbf{Y}=\mathcal{F}_{conv}(\mathbf{X}) + \mathbf{X}$. 
The shortcut structure facilitates gradient backpropagation and encourages residual learning.
DenseNet~\cite{huang2016densely} instead proposes direct feature reuse by forwarding and appending input channels $\mathbf{X}$ directly to $\mathbf{Y}$, specifically $\widetilde{\mathbf{Y}} = [\mathbf{X}; \mathbf{Y}]$.

\subsubsection{Drop-neuron - The Neuron Level Dropout}

Dropout~\cite{srivastava2014dropout,hinton2012improving} is widely adopted in the training of deep neural networks as an effective regularization and implicit model ensemble method.
The standard dropout is applied to each input neuron with a single parameter $p$ during training, controlling the participation of each neuron $x_j$ with a gating variable $\alpha_j$ for each forward pass:

\begin{equation}\label{formular:drop-neuron}
y_i = \frac{1}{p} \sum_{j=1}^{N} w_{ij} (\alpha_j \cdot x_j), \alpha_j \sim \mathcal{B}ernoulli(p)
\end{equation}

where $\cdot$ denotes scalar multiplication and $\alpha_j$ is an \textit{independent} Bernoulli random variable which takes the value 1 with probability $p$ (the retain ratio) and the value 0 with probability $q = 1-p$ (the drop ratio).
The scaling factor $\frac{1}{p}$ scales the output activation to keep the expected value of the output during training.
Then during inference, the transformation is simply the same as Equation~\ref{formular:neuron_op}.
We name the neuron level dropout \dropneuron to distinguish it from other higher structural levels of dropout variants.
\Dropneuron introduces randomness to the training process, which forces each neuron to learn more robust representations that are effective with varying input neuron set, and thus improves generalization.
Therefore, the resultant network for inference can be regarded as an exponentially-sized ensemble of all possible subnets.

As illustrated in the operation 1 of Figure~\ref{fig:conv_architecture}, \dropneuron empirically proves to be effective for deep neural networks, especially for dense layers such as fully-connected layers and recurrent layers.
However, recent CNN architectures~\cite{he2016deep,huang2016densely,hu2017squeeze} find that dropout is ineffective for convolutional layers.
We note that this can be attributed to the fact that in convolutional layers, neurons within the same input channel are highly spatial correlated, and features are extracted channel-wise.
Therefore, dropping neurons independently can hardly regularize the training of deep CNNs, whose effectiveness is further reduced when CNNs train with extensive data augmentation.

\subsubsection{\Dropchannel - The Channel Level Dropout}
\label{sec:drop_channel}


The channel level dropout, i.e., \dropchannel, is inspired by the observation that there exists a close structural correspondence between channels in the convolutional layer and neurons in canonical neural networks, which is formulated formally in Equation~\ref{formular:neuron_op} and Equation~\ref{formular:conv_op}.
Therefore, dropping channels should be structurally more effective in regularization.
Following a similar methodology, \dropchannel can be formulated as:

\begin{equation}\label{formular:drop-channel}
\mathbf{y}_i = \mathbf{w}_i \ast \mathbf{\widetilde{X}} = \frac{1}{p} \sum_{j=1}^{C_{in}} \mathbf{w}_{i}^{j} \ast (\alpha_j \cdot \mathbf{x}_j)
\end{equation}

where $\alpha_j$ is again an \textit{independent} Bernoulli random variable with probability $1-p$ of being 1 and is applied to the entire channel $\mathbf{x}_j$.
Particularly, $\alpha_j$ controls the presence of the input channel $x_j$ during training.
The \dropchannel training is illustrated in Figure~\ref{fig:conv_architecture}, where the identity mapping of operation 2 is replaced by operation 3, a random sampling of $\alpha_j$ followed by the corresponding gating on input channels.
The scaling factor $\frac{1}{p}$ here again compensates for the scale loss from the deactivation of input channels and keeps the expected value of the output during training.
The resultant network after training can then be directly used for inference with Equation~\ref{formular:conv_op}.

The idea of channel level dropout is first introduced in SpatialDropout~\cite{tompson2015efficient}.
However, \cite{tompson2015efficient} only shows that SpatialDropout improves CNN models over an object localization dataset, and the effectiveness of SpatialDropout in interaction with other training techniques, e.g., data augmentation and batch normalization~\cite{ioffe2015batch}, are not investigated.
To better harness the regularization and ensemble effects, we examine the complex interaction between \dropchannel and these techniques extensively used in state-of-the-art CNNs theoretically and empirically.

\begin{equation}
\label{formular:batch_norm}
    \mathbf{\hat{x}_j} = \frac{\mathbf{x}_{j}-\mu_j}{\sqrt{\sigma_j^2+\epsilon }}; \mathbf{x}_{j}'=\gamma_j \mathbf{\hat{x}}_j + \beta_j 
\end{equation}

\begin{figure}[t]
    \centering
    \subfloat[traditional]{
        \includegraphics[width=0.14\textwidth]{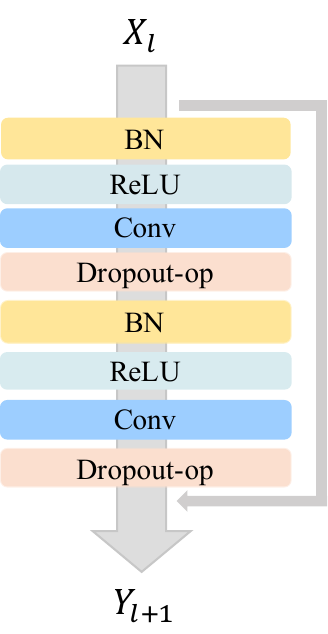}
        \label{fig:traditional}
    }
    \hspace{0.04\textwidth}
    \subfloat[proposed]{
        \includegraphics[width=0.14\textwidth]{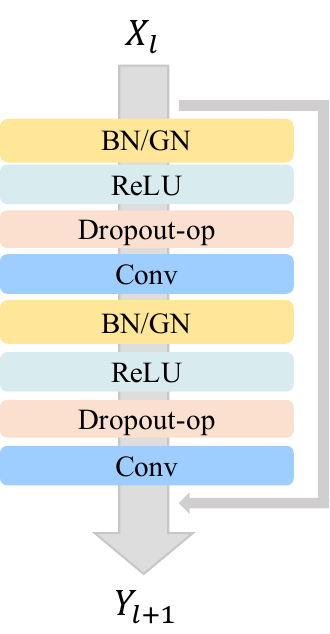}
        \label{fig:proposed}
    }
    \caption{The convolutional building blocks with drop-operations, for both \dropneuron and \dropchannel, demonstrated with the pre-activation residual convolution block.}
    \label{fig:conv_formulation}
\end{figure}

Each convolutional layer of deep CNNs is typically followed with a batch normalization layer (BN)~\cite{ioffe2015batch} to normalize inputs batch-wise, which stabilizes the mean and variance of the input channels $\mathbf{X}$ received by each output channel $\mathbf{y}_i$.
Take the pre-activation convolutional layers~\cite{he2016identity,huang2016densely,hu2017squeeze} for example, the convolutional transformation follows the \textit{BN-ReLU-Conv} transformation pipeline, as illustrated in Figure~\ref{fig:traditional}.
We note that the dropout operation, including \dropneuron and \dropchannel, is not integrated into the convolutional blocks properly, which is either totally discarded in recent CNNs or used in an erroneous way.
As shown in Equation~\ref{formular:batch_norm}, the BN layer normalizes each input channel $\mathbf{x}_{j}$ with the batch channel mean $\mu_j$ and variance $\sigma_j$, and keeps running estimates of them, which will be directly used for the normalization of the $j_{th}$ input channel after training.
$\gamma_j$ and $\beta_j$ are learnable affine transformation parameters associated with channel $\mathbf{x}_{j}$, and $\epsilon$ is for computation stability.

However, the two dropout operations are traditionally introduced right between the convolutional layer and the BN layer, which leads to violent fluctuation of the mean and variance of inputs received by the BN layer, for both \dropneuron and \dropchannel.
We attribute the failure of the traditional \dropneuron and \dropchannel to the incorrect placement of the dropout operations and propose general convolutional building blocks with the dropout operations placed right before each convolutional layer in Figure~\ref{fig:proposed}.
Integrating drop-operations before the convolutional operation, as is also analyzed in the local reparameterization of the variational dropout~\cite{kingma2015variational}, leads to lower gradient variance and thus faster convergence.
In Section~\ref{sec:experiment}, we will also validate empirically the effectiveness of the proposed convolutional blocks on various state-of-the-art CNNs with extensive experiments.

The disharmony between \dropneuron and the BN layer has also been analyzed in~\cite{li2019understanding}, which shows that there exists inconsistency of the variance estimation of BN from training to inference when the BN layer coupled with the neuron level dropout.
The variance shift leads to unstable predictions and thus probably worse inference performance.
A simple way to reduce this issue is to update the variance estimation~\cite{adjustingBN} of BN layers after training without dropout.
In this paper, we further formally examine \dropchannel and show that for CNNs, variance shift can be largely reduced with dropout operations placing right before the convolutional layer.
In CNNs, inputs are normalized channel-wise, and different channels transform with different filters.
Particularly, each channel $\mathbf{x}_{j}$ of the input $\mathbf{X}$ is obtained by convolution over inputs of its preceding layer with an independent filter, whose mean and variance are then subsequently normalized with BN independently of other channels.
Therefore,  we can denote that the $j_{th}$ input channel $\mathbf{x}_{j}$ shares the same mean $\beta_j$ and variance $\gamma_j^2$ (i.e. $\E(x_{j,\cdot})=\beta_j$ and $\Var(x_{j,\cdot})=\gamma_j^2$), and assume uncorrelatedness between different channels (i.e. $\Cov(x_{i, \cdot}, x_{j,\cdot})=0, i \ne j $)~\footnote{In practice, obtaining fully uncorrelatedness requires expensive input whitening on the input $\mathbf{X}$.}.
Then for \dropchannel in the traditional block, dropout operates on $\mathbf{y}_{i}$ as in Equation~\ref{formular:conv_op} (the input $\mathbf{x}_{i}$ of the next layer), and outputs $y_{i,k}' = \frac{1}{p}\alpha_i y_{i,k}$ during training.
Denoting $\E(\mathbf{y}_{i,\cdot})=\beta_i$ and $\Var(\mathbf{y}_{i,\cdot})=\gamma_i^2$, and omitting subscript $i,k$, we have: 

\begin{equation} \label{formular:var_traditional}
\begin{split}
    \Var^{Train}(y') &= \frac{1}{p^2} \E(\alpha^2) \E(y^2) - \frac{1}{p^2}(\E(\alpha) \E(y))^2 \\
                     &= \frac{1}{p} \gamma^2 + \frac{1-p}{p} \beta^2
\end{split}
\end{equation}

Conventionally, the BN layer following $\mathbf{y}_{i}$ keeps a record of the variance $\Var^{Train}(y_i')$ and uses the running estimate of it during inference.
This variance estimation deviates from the actual variance $\Var^{Test}(y') = \gamma^2$, and the shift ratio is:

\begin{equation} \label{formular:shift_traditional}
    \varDelta(p)' = \frac{\Var^{Train}(y')}{\Var^{Test}(y')} = \frac{1}{p} + \frac{1-p}{p}\frac{\beta^2}{\gamma^2}
\end{equation}

Therefore, there exists a variance shift of ratio $\varDelta(p) \ge 1$, and $\varDelta(p) = 1$ only when the channel retain ratio $p$ is 1, i.e., the absence of dropout operation.
In the proposed convolutional building block, dropout instead operates on $\mathbf{x}_{i}$ as in Equation~\ref{formular:drop-channel} right after the convolution operation.
We first vectorize the filter for each input channel $\mathbf{x}_{i}$;
then for each output $y_{i, k}$ of $\mathbf{y}_{i}$, we have $y_{i, k} = \frac{1}{p} \sum_{j=1}^{C_{in}} \sum_{d=1}^{D} w_{i,d}^{j} \cdot (\alpha_j \cdot x_{j,k \ast d})$, where $x_{j,k \ast \cdot}$ denotes the vectorized receptive filed of $y_{i,k}$. Omitting the subscript $i,k$ and with uncorrelatedness we have:

\begin{equation} \label{formular:var_proposed_train}
\begin{split}
    \Var^{Train}(y) 
    &= \frac{1}{p^2} \sum_{j=1}^{C_{in}} \Var [\alpha_j \sum_{d=1}^{D} (w_{d}^{j} x_{j, d} )] \\
    &= \frac{1}{p} \Var^{Test}(y) + \frac{1-p}{p}\sum_{j=1}^{C_{in}}\beta_j^2 (\sum_{d=1}^{D} w_{d}^{j})^2
\end{split}
\end{equation}

where the inference variance is:

\begin{equation} \label{formular:var_proposed_test}
\begin{split}
    \Var^{Test}(y) &= \sum_{j=1}^{C_{in}}\Var( \sum_{d=1}^{D} (w_{d}^{j} x_{j, d} ) \\
    &= \sum_{j=1}^{C_{in}} \gamma_j^2 \sum_{m=1}^D \sum_{n=1}^D w_{m}^{j} w_{n}^{j} \rho^j_{m,n}
\end{split}
\end{equation}

where $\rho_{m, n}^j = \frac{\Cov(x_{j,m}, x_{j,n})}{\sqrt{\Var(x_{j,m})} \sqrt{\Var(x_{j,n})} } \in [-1, 1]$ measures the linear correlation between $x_{j,m}$ and $x_{j,n}$.
Therefore, the shift ratio of the proposed \dropchannel block is:

\begin{equation} \label{formular:shift_proposed}
\begin{split}
    \varDelta(p) &= \frac{\Var^{Train}(y)}{\Var^{Test}(y)} \\
    &= \frac{1}{p} + \frac{1-p}{p} \frac{\sum_{j=1}^{C_{in}} \beta_j^2 \sum_{m=1}^D \sum_{n=1}^D w_{m}^j w_{n}^j} {\sum_{j=1}^{C_{in}} \gamma_j^2 \sum_{m=1}^D \sum_{n=1}^D w_{m}^j w_{n}^j \rho_{m,n}^{j}}
\end{split}
\end{equation}

We note that although $\varDelta(p) \ge 1$, the variance shift of \dropchannel after the convolution is smaller and more stable than $\varDelta(p)'$.
For $\varDelta(p)'$, the $\beta_j$ and $\gamma_j^2$ are the mean and variance of the channel activations after the convolution, which keeps evolving during the training process and could be unbounded.
In contrast, for $\varDelta(p)$, $\beta_j$ and $\gamma_j^2$ are typically the mean and variance of the channel activations normalized by its preceding BN layer and is thus stable; and the kernel weight $w_{i,d}^j$ and the correlation between input $x_{j, m}$ and $x_{j, n}$ in the same channel $\mathbf{x}_{j}$ are also relatively stable as the training progresses.

Empirically, we observe less variance shift with \dropchannel placed right before the convolutional layer and with a relatively large channel retain ratio (e.g., 0.9), whose shift ratio is rather close to BN training without dropout.
We note that the variance shift of \dropchannel can be greatly reduced by updating BN running estimates after training, and alternatively, the regularization and implicit model ensemble benefits from \dropchannel can be fully harnessed by replacing BN with Group Normalization~\cite{wu2018group} (GN).
GN normalizes channels within the same channel group of each layer instead of batch-wise as in BN, and requires no running estimates of the channel mean and variance and thus leads to no variance shift.
Compared with \dropneuron, CNNs trained with our proposed \dropchannel generally improve training considerably.

\subsubsection{Higher Level Dropouts: Drop-paths}
\label{sec:higher_level_dropout}

The path level dropout \droppath are introduced in FractalNet~\cite{larsson2016fractalnet}.
Notably, ResNet with Stochastic Depth~\cite{huang2016deep} proposes to randomly drop the residual path and forward the input $\mathbf{X}$ during training with the shortcut connection of operation 5, which is denoted as \droplayer and can be regarded as a \droppath that randomly drops the transformation path.
Formally, \droppath can be formulated as:

\begin{equation}\label{formular:drop-path}
\mathbf{Y} = \mathcal{F}_{conv}(\mathbf{X}) = \frac{1}{p} \sum_{i=1}^{P} \alpha_i \cdot \mathcal{F}_{conv_\_i}(\mathbf{X})
\end{equation}

where $P$ is the number of paths (branches) of the convolutional layer, and $\alpha_i$ is a Bernoulli gating variable that controls the participation of $i_{th}$ path in the transformation with the retain ratio $p$.
Although \droppath and \droplayer are effective in regularizing CNNs, they are highly dependent on CNN architectures.
Specifically, \droppath requires CNN to contain multiple paths, either homogeneous or heterogeneous, of $\mathcal{F}_{conv}$ in operation 4; and particularly, \droplayer demands shortcut connection of operation 5 illustrated in Figure~\ref{fig:conv_architecture}.
In FractalNet, \droppath is applied to the fractal architecture, whose paths are heterogeneous.

We note that \droppath can be integrated as a general building block with the bottleneck structure~\cite{he2016deep} and group convolution~\cite{krizhevsky2012imagenet}.
Specifically, the building block is based on the bottleneck structure of one 3$\times$3 convolution surrounded by dimensionality reducing and expanding with a 1$\times$1 convolution, i.e., \textit{conv1$\times$1-conv3$\times$3-conv1$\times$1}.
To support \droppath, group convolution is introduced to the inner 3$\times$3 convolutional layer with $P$ groups as proposed by ResNeXt~\cite{xie2017aggregated}.
Then structurally, the bottleneck building block contains $P$ independent paths of homogeneous transformations, each of which first collapses $C$ input channels into $d$ channels by a 1$\times$1 convolution, and then transforms by an inner 3$\times$3 convolution within each path and finally expands back to $C$ channels by a 1$\times$1 convolution.
In this implementation, such a building block is equivalent to the bottleneck with the inner 3$\times$3 convolution of $P$ groups.
We further propose to choose $P$ to be a power of 2 empirically, e.g., 16, 32, 64 and fix $P$ for the whole network, and $d = \frac{C}{2 P}$.
Further, we find that \droplayer is better in line with the shortcut connection of identity mapping~\cite{he2016identity,zagoruyko2016wide}, i.e., $\mathbf{Y} = \mathcal{F}_{conv}(\mathbf{X}) + \mathbf{X}$, where output scaling is not needed since the residual path $\mathcal{F}_{conv}(\mathbf{X})$ for random dropping is pushed towards zero~\cite{he2016deep}.

\subsubsection{Dropout as General Training Regularization for CNNs}
\label{general_mechanisms}

Thus far, we have formulated and examined three different structural levels of dropout, i.e., \dropneuron, \dropchannel and \droppath.
The effectiveness of different dropout variants can be mainly attributed to the regularization and ensemble effect~\cite{hinton2012improving,wager2013dropout,srivastava2014dropout}.
We note that \dropneuron and \dropchannel are readily applicable to existing CNNs with the adjustment of placing the dropout operation right before each convolutional transformation; we have also proposed corresponding building blocks with \dropneuron/\dropchannel in Section~\ref{sec:drop_channel} and \droppath in Section~\ref{sec:higher_level_dropout}.

Inspired by \dropchannel (Equation~\ref{formular:drop-channel}) and \droppath (Equation~\ref{formular:drop-path}), we further propose \Dropout for CNNs as a general regularization technique, which treats each channel connection between input and output channels as a channel path and then replicates each path $P$ times:

\begin{equation}\label{formular:drop-conv2d}
\begin{aligned}
\mathbf{y}_i = \mathbf{\widetilde{w}}_i \ast \mathbf{X} &= \sum_{j=1}^{C_{in}} (\frac{1}{p} \alpha_j \cdot \mathbf{w}_{i}^{j})\ast \mathbf{x}_j \\
&=  \sum_{j=1}^{C_{in}} (\sum_{k=1}^{P} \frac{1}{p} \alpha_{j,k} \cdot \mathbf{w}_{i}^{j,k})\ast \mathbf{x}_j
\end{aligned}
\end{equation}

where $P$ is the number of path replicates for each channel connection, and $\alpha_{j,k}$ is again an independently sampled Bernoulli gating variable.
In implementation, Equation~\ref{formular:drop-channel} of \dropchannel can be equivalently regarded as convolution over each input channel $\mathbf{x}_j$ with dropout weight $\mathbf{\widetilde{w}}_i^j = \frac{1}{p} \alpha_j \cdot \mathbf{w}_{i}^{j}$, where the coefficient $\frac{1}{p} \alpha_j$ is the scaling factor during training.
For \dropout, we enhance each channel path with an ensemble of $P$ paths of different weights during training, i.e. $\mathbf{w}_i^j = \sum_{k=1}^{P} \mathbf{w}_{i}^{j,k}$, and scale each weight $\mathbf{w}_{i}^{j,k}$ with dropout scaling factor $\frac{1}{p} \alpha_{j,k}$ accordingly.

We note that the \dropout enhancement leads to both larger model capacity and better dropout training, harnessing regularization and model ensemble benefits for improved training.
More importantly, such enhancement incurs negligible costs.
Specifically, during training, we only need to replicate the weights $P$ times, and the convolution of inputs with these weights can be equivalently implemented by first a weighted average of these weights and then the computational heavy convolution.
Then during inference, the $P$ channel weights can be aggregated back into one, and the convolution with \dropout is therefore the same as the one trained without \dropout.
We further note that these dropout variants (particularly \dropout) can be readily integrated into deep CNNs by simply replacing the convolutional blocks with our proposed ones and then configuring the dropout rate for the required regularization strength, which generally takes no additional model parameter and incurs negligible computational cost while effectively improve the training of CNNs.
Further, these dropout variants can be jointly adopted in CNNs whenever possible, e.g., \dropchannel, \droppath can be adopted in ResNeXt~\cite{xie2017aggregated} concurrently.

\section{Experiments}
\label{sec:experiment}

\begin{table*}[ht]
    \centering
    \renewcommand{\arraystretch}{.9}
    \resizebox{1.8\columnwidth}{!}{
        \begin{tabular}{ c||c|cccc|c}
    
    \thickhline
    Group   &   Output Size &   VGG-11   &   WRN-16-8 &   ResNeXt-29-P64-d4 & DenseNet-L190-K40 &   WRN-40-4 \\
    \hline
    conv1   &   32$\times$32  &   [conv3$\times$3, 64]$\times$2  &   [conv3$\times$3, 16]$\times$1   &   [conv3$\times$3, 64]$\times1$   &   [conv3$\times$3, 80]$\times1$   &   [conv3$\times$3, 16]$\times1$\\
    conv2   &   32$\times$32  &  -  &   [Block, 16$\times$4]$\times$6   &   [B-Block, 256]$\times4$       &   [D-Block, 80-1320]$\times31$ &   [Block, 16$\times8$]$\times2$\\
    conv3   &   16$\times$16  &   [conv3$\times$3, 256]$\times$2  &   [Block, 32$\times$4]$\times$6  &   [B-Block, 512]$\times4$     &   [D-Block, 660-1900]$\times31$  &   [Block, 32$\times8$]$\times2$\\
    conv4   &   8$\times$8  &   [conv3$\times$3, 256]$\times$2  &   [Block, 64$\times$4]$\times$6 &   [B-Block, 1024]$\times4$       &   [D-Block, 950-2190]$\times31$   &   [Block, 64$\times8$]$\times2$\\
    conv5   &   8$\times$8  &   [conv3$\times$3, 512]$\times$4  &    -   &   -&   -       &   -\\
    avgPool    &   1$\times$1  &   [avg8$\times$8, 512] &   [avg8$\times$8, 256]&   [avg8$\times$8, 1024]    &   [avg8$\times$8, 2190] &   [avg8$\times$8, 512]\\
    \hline
    Dataset & - & CIFAR & CIFAR & CIFAR & CIFAR & SVHN\\
    Params & - & 9.89M & 8.95M & 8.85M & 25.62M & 10.96M\\
    \thickhline
        \end{tabular}
    }
    \caption{
    Detailed architectures and configurations of representative convolutional neural networks for CIFAR and SVHN datasets. Building blocks are denoted as ``[block, number of channels] $\times$ number of blocks''. 
    }
    \label{tab:cnns_cifar}
\end{table*}

The four structural levels of dropouts are evaluated on representative CNNs on widely benchmarked datasets, including CIFAR, SVHN and ImageNet.
We first introduce the dataset and training details and CNN architectures.
We then evaluate the effectiveness of our building blocks with the proposed dropout operations.
We compare \dropneuron, \dropchannel, \droppath (also \droplayer), \dropout and their combinations, with which we improve over state-of-the-art CNNs on CIFAR and SVHN datasets and achieve consistently better results.

\subsection{Dataset Details}

\subsubsection{CIFAR}

The two CIFAR~\cite{krizhevsky2009learning} datasets consist of $32 \times 32$ colored scenery images. 
CIFAR-10 (C10) consists of images drawn from 10 classes, and CIFAR-100 (C100) from 100 classes.
The training and testing set for both datasets contain $50,000$ and $10,000$ images respectively.
Following the standard data augmentation scheme~\cite{he2016deep,huang2016deep,huang2016densely}, 
each image is first zero-padded with 4 pixels on each side, 
then randomly cropped to produce $32 \times 32$ images again, followed by a random horizontal flip. 
We denote the datasets with data augmentation with ``+'' behind the dataset names (e.g., C10+). 
We normalize the data using the channel means and standard deviations for data preprocessing.

\subsubsection{SVHN}

The Street View House Numbers dataset~\cite{netzer2011reading} contains $32 \times 32$ colors digit images from Google Street View. 
The task is to correctly classify the central digit into one of the 10 digit classes. 
The training and testing sets respectively contain $73,257$ and $26,032$ images, and an additional training dataset contains $531,131$ images that are relatively easier to classify.
We adopt a common practice~\cite{huang2016deep,zagoruyko2016wide,huang2016densely} by using all the training data without any data augmentation. 
Following~\cite{zagoruyko2016wide,huang2016densely}, we divide each pixel value by 255, scaling the input to range $[0,1]$.

\subsubsection{ImageNet}
The ILSVRC 2012 image classification dataset contains 1.2 million images for training and 50,000 for validation from 1000 classes.
We adopt the same data augmentation scheme for training images following the convention~\cite{he2016deep,zagoruyko2016wide,huang2016densely}, and apply a $224\times224$ center crop to images at test time.
The results are reported on the validation set.

\subsection{CNN Architecture Details}
\label{cnn_architecture_detail}

As discussed in Section~\ref{general_mechanisms}, \dropneuron and \dropchannel are generally applicable to CNNs, while the applicability of \droppath and \droplayer are dependent on CNN architectures.
To evaluate the effectiveness of these dropout variants, we adopt CNN architectures with representative convolutional transformation.

Specifically, we first evaluate and compare \dropneuron and \dropchannel with our proposed building blocks illustrated in Figure~\ref{fig:proposed} on VGG~\cite{simonyan2014very}, whose convolutional layer is a plain 3$\times$3 conv following operation 2 and 4 of Figure~\ref{fig:conv_architecture}.
For \droppath of the new building block proposed in Section~\ref{sec:higher_level_dropout} and \droplayer, we validate their effectiveness in comparison with \dropchannel and \dropneuron on ResNeXt~\cite{xie2017aggregated} with (1 multiple paths and residual connection, (2 Wide Residual Networks~\cite{zagoruyko2016wide} (WRN) with wider convolutional layer and residual connection and (3 DenseNet~\cite{huang2016densely} with shortcut connection.

We denote these models with WRN-depth-k, ResNeXt-depth-P-d and DenseNet-L-K, with k, P, d, L, K as the widening factor~\cite{zagoruyko2016wide}, the number of paths, the channel width of each path~\cite{xie2017aggregated}, the number of layers (depth) and K the growth rate~\cite{huang2016densely}, respectively.
The building blocks of WRN, ResNeXt and DenseNet are basic block (Block) with two consecutive 3$\times$3 conv, the bottleneck block (B-Block) proposed in Section~\ref{sec:higher_level_dropout} and the DenseNet bottleneck block (D-Block)~\cite{huang2016densely} with dimensionality reduction of 1$\times$1 conv and the following transformation of 3$\times$3 conv.
The detailed model configurations for CIFAR-10/100, SVHN and ImageNet datasets are introduced in Table~\ref{tab:cnns_cifar} and Table~\ref{tab:cnns_imagenet}.

For CNNs trained with \dropneuron and \dropchannel, each convolutional layer is replaced with the proposed building block in Section~\ref{dropout_for_cnn}, where the dropout operations are integrated into the transformation right before the convolution.
While for \droppath, each convolutional layer is replaced with the proposed general \droppath building block in Section~\ref{sec:higher_level_dropout}.

\begin{table*}[ht]
    \centering
    \resizebox{1.6\columnwidth}{!}{
        \begin{tabular}{ c||c|c||c|cccc}
    
    \thickhline
    Group   &   Output Size &   VGG-16 & Output Size & WRN-50-2 & DenseNet-L169-K32\\
    \hline
    conv1   &   224$\times$224  &   [conv3$\times$3, 64]$\times$2  &   112$\times$112 & conv7$\times$7, stride 2    &   conv7$\times$7, stride 2 \\
    pooling & 112$\times$112 & 2$\times$2 max pool, stride 2 & 56$\times$56 & 3$\times$3 max pool, stride 2 & 3$\times$3 max pool, stride 2\\
    
    conv2   &   112$\times$112  &  [conv3$\times$3, 128]$\times$2  & 56$\times$56   &   [B-Block, 64$\times$2]$\times$6  &   [D-Block, 64-256]$\times$6\\
    pooling & 56$\times$56 & 2$\times$2 max pool, stride 2 & 28$\times$28 & conv3$\times$3, stride 2 & Transition Layer \\
    
    conv3   &   56$\times$56  &  [conv3$\times$3, 256]$\times$2  & 28$\times$28   &   [B-Block, 128$\times$2]$\times$3  &   [D-Block, 128-512]$\times$12\\
    pooling & 28$\times$28 & 2$\times$2 max pool, stride 2 & 14$\times$14 & conv3$\times$3, stride 2 & Transition Layer \\
    
    conv4   &   28$\times$28  &  [conv3$\times$3, 512]$\times$2  & 14$\times$14   &   [B-Block, 256$\times$2]$\times$3  &   [D-Block, 256-1792]$\times$32\\
    pooling & 14$\times$14 & 2$\times$2 max pool, stride 2 & 7$\times$7 & conv3$\times$3, stride 2 & Transition Layer \\
    
    conv5   &   14$\times$14  &  [conv3$\times$3, 512]$\times$2  & 7$\times$7   &   [B-Block, 512$\times$2]$\times$3  &   [D-Block, 896-1920]$\times$32\\
    pooling & 7$\times$7 & 2$\times$2 max pool, stride 2 & 1$\times$1 & 7$\times$7, global avg-pool & 7$\times$7, global avg-pool \\

    FC    &   1000  &   [512$\times$7$\times$7,4096,4096,1000] &   1000 & [1024,1000] &   [1920,1000]\\
    \hline
    Params & - & 138.4M & - & 68.9M & 14.1M\\
    \thickhline
        \end{tabular}
    }
    \caption{Detailed configurations of representative CNNs for the ImageNet dataset.
    }
    \label{tab:cnns_imagenet}
\end{table*}

\subsection{Training Details}

For all the experiments, we train the CNNs with SGD and Nesterov momentum.
For CIFAR datasets, we train 300 epochs for VGG-11, ResNeXt-29-P64-d4 and DenseNet-L190-K40, and 200 epochs for WRN-40-4.
For SVHN, we train 160 epochs for WRN-16-8.
The initial learning rate is set to 0.1, weight decay 0.0001, dampening 0, momentum 0.9 and mini-batch size 128 for CIFAR and SVHN datasets.
The learning rate is divided by 10 at 50\% and 75\% of the total number of training epochs.
For ImageNet, we train 100 epochs for VGG-16, WRN-50-2 and DenseNet-L169-K32 with a mini-batch size of 128.
The initial learning rate is set to 0.05, and is lowered by a factor of 10 after epoch 30, 60 and 90.
We use a weight decay of 0.0001 and momentum 0.9 without dampening.

\subsection{Experimental Results}

\subsubsection{Building Blocks of \Dropneuron and \Dropchannel}

\begin{table}[h!]
\centering
\scalebox{0.94}{
    \begin{tabular}{ c || c | c  c | c  c }
    
    \thickhline
    \multicolumn{1}{c||}{Network} & {original} & {DN} & {DN+} & {DC} & {DC+} \\
    \hline
    
    \multicolumn{1}{c||}{VGG} & {\small 5.09} & {\small 5.18} & {\small 4.98 (+0.20)} & {\small 4.78} & {\small \textbf{4.67} (+0.11)} \\
    {WRN} & {\small 4.97} & {\small 4.89} & {\small 4.63 (+0.26)} & {\small 4.60} & {\small \textbf{4.31} (+0.29)} \\
    {DenseNet} & {\small 4.57} & {\small 4.70} & {\small 4.52 (+0.18)} & {\small 4.56} & {\small \textbf{4.32} (+0.24)} \\
    
    \thickhline
    \end{tabular}
}
\caption{Error rates (\%) of CNNs trained with dropout operations w/ and w/o the proposed blocks on CIFAR-10.}\label{tab:drop_operation_comp}
\end{table}

We validate the effectiveness of the proposed building blocks supporting \dropneuron and \dropchannel on VGG-11, WRN-40-4 and DenseNet-L100-K12, as introduced in Section~\ref{cnn_architecture_detail} and Table~\ref{tab:cnns_cifar}.
The results are summarized in Table~\ref{tab:drop_operation_comp}, where we report the results of networks trained without dropout (original), with traditional \dropneuron and \dropchannel (DN and DC), and with the proposed \dropneuron and \dropchannel (DN+ and DC+) respectively.

\begin{figure}[t]
    \centering
    \includegraphics[width=0.42\textwidth]{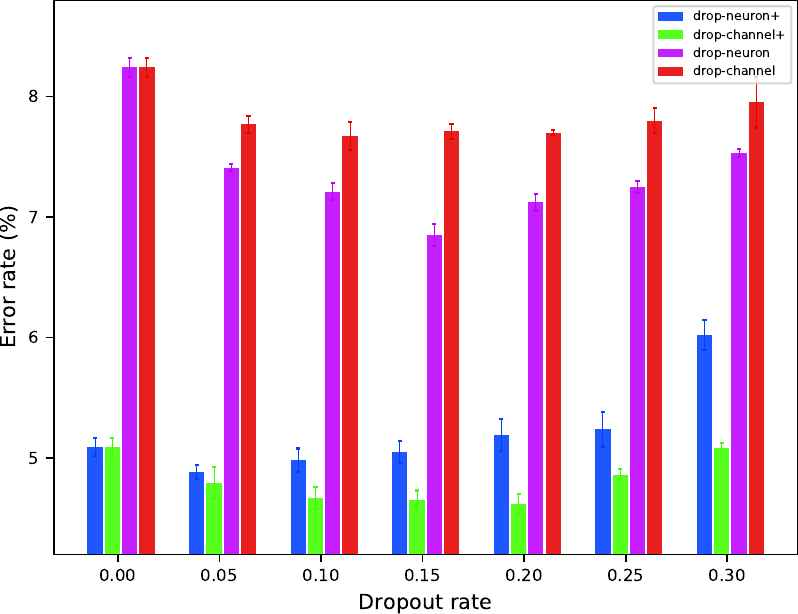}
    \caption{Error rate (\%) of VGG-11 trained with \dropneuron and \dropchannel w/ and w/o data augmentation.}
    \label{fig:vgg_10_bars}
\end{figure}

We can notice that the proposed building blocks consistently improve over the original blocks by a significant margin.
Comparing to the original results, networks trained with \dropneuron and \dropchannel achieve significantly better performance, which demonstrates that the dropout technique is effective in regularizing CNNs if applied properly.
For instance, the introduction of \dropchannel alone achieves a reduction of the error rate by 0.42\%, 0.66\% and 0.25\% on VGG-11, WRN-40-4 and DenseNet-L100-K12 respectively.

\subsubsection{Dropout operations and Data Augmentation}
\label{data_augmentation}

\begin{figure}[t]
    \centering
    \includegraphics[width=0.42\textwidth]{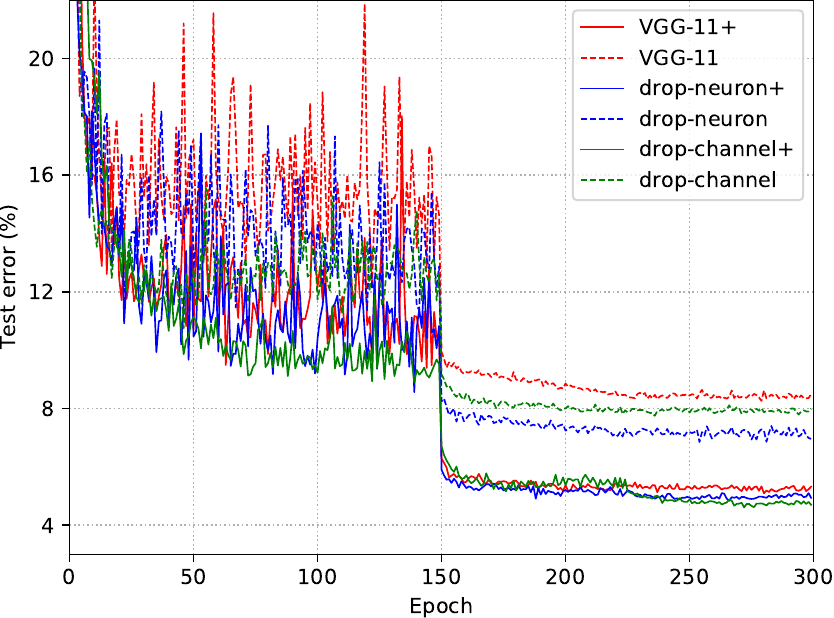}
    \caption{Learning curves of VGG-11 trained w/ and w/o \dropneuron, \dropchannel and data augmentation.}
    \label{fig:vgg_curve_all}
\end{figure}

We evaluate the relationship between data augmentation and dropout operations of drop-neuron and drop-channel for CNNs.
The results are reported on VGG-11, whose error rates and learning curves are illustrated in Figure~\ref{fig:vgg_10_bars} and Figure~\ref{fig:vgg_curve_all}.
We denote VGG networks trained without dropout, with drop-neuron and drop-channel as VGG-11, drop-neuron and drop-channel respectively, and the network trained with standard data augmentation is marked with a suffix $\mathbf{+}$.

We summarize the main results in Figure~\ref{fig:vgg_10_bars}, where error rates and standard deviations are reported with the dropout rate in every 0.05.
The results show that data augmentation is essential for CNNs; without data augmentation, the performance decreases by around 3\%.
Further, \dropneuron and \dropchannel improve the performance noticeably both with and without data augmentation.
With data augmentation and \dropchannel, the model achieves the best result of 4.62\% from 8.24\%, i.e., a 3.62\% error rate reduction; meanwhile without data augmentation, \dropneuron achieves a better result of 6.85\% than \dropchannel.
These results show that the regularization effect of \dropneuron overlaps with data augmentation to some extent.

We further plot learning curves of CNNs trained with the best dropout rates in Figure~\ref{fig:vgg_curve_all}.
The learning curves confirm these findings and demonstrate that \dropneuron and \dropchannel are indeed effective in regularizing CNNs.
For state-of-the-art CNNs trained with extensive data augmentation, \dropchannel is more effective in regularization, which improves the performance by a noticeable margin.

\subsubsection{The effectiveness of \Droppath and \Droplayer}

We evaluate \droppath on ResNeXt-29-64-4 (see Table~\ref{tab:cnns_cifar}), specifically the effect of \droppath alone and with other dropout operations.
We adopt building blocks supporting \droppath as proposed in Section~\ref{sec:higher_level_dropout}.

\begin{figure}[t]
    \centering
    \includegraphics[width=0.42\textwidth]{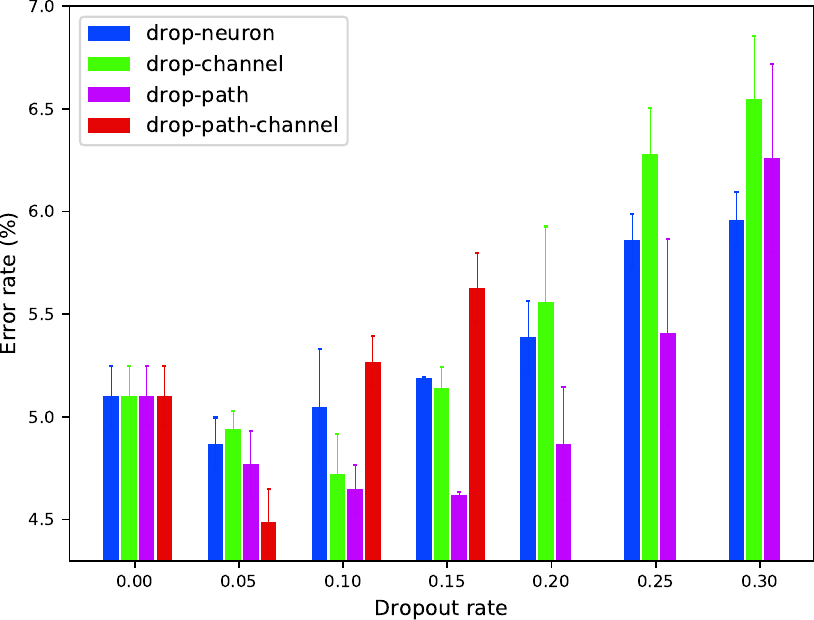}
    \caption{Error rate (\%) of ResNeXt-29-64-4 trained with drop-neuron, drop-channel and drop-path.}
    \label{fig:resnext_10_bars}
\end{figure}

The results of ResNeXt trained with \dropneuron, \dropchannel, \droppath and dropout with both \droppath and \dropchannel (drop-path-channel, with the same dropout rate from 0.05 to 0.15) are summarized in Figure~\ref{fig:resnext_10_bars}.
Results show that \dropneuron reduces the error rate modestly from 5.10\% to 4.87\%, and \dropchannel outperforms \dropneuron with a lower error rate of 4.72\%.
With the proposed \droppath building block, ResNeXt achieves a better result of 4.62\%, i.e., a 0.48\% relative reduction over the network without dropout.

As discussed in Section~\ref{general_mechanisms}, the path level dropout can be adopted with finer-grained dropout operations, i.e., \dropneuron and \dropchannel.
With both \droppath and \dropchannel, ResNeXt achieves the lowest error rate of 4.49 with a dropout rate of 0.05.
This result confirms that different structural levels of dropouts can further improve the training of CNNs.

\begin{figure}[t]
    \centering
    \includegraphics[width=0.42\textwidth]{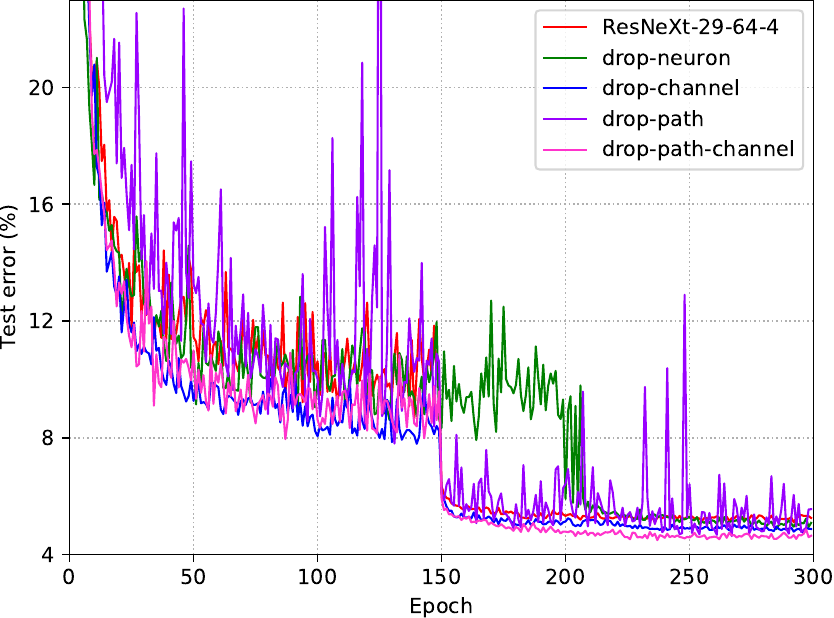}
    \caption{Learning curves of ResNeXt-29-64-4 trained with \dropneuron, \dropchannel and \droppath.}
    \label{fig:resnext_curve_all}
\end{figure}

\begin{figure}[t]
    \centering
    \includegraphics[width=0.42\textwidth]{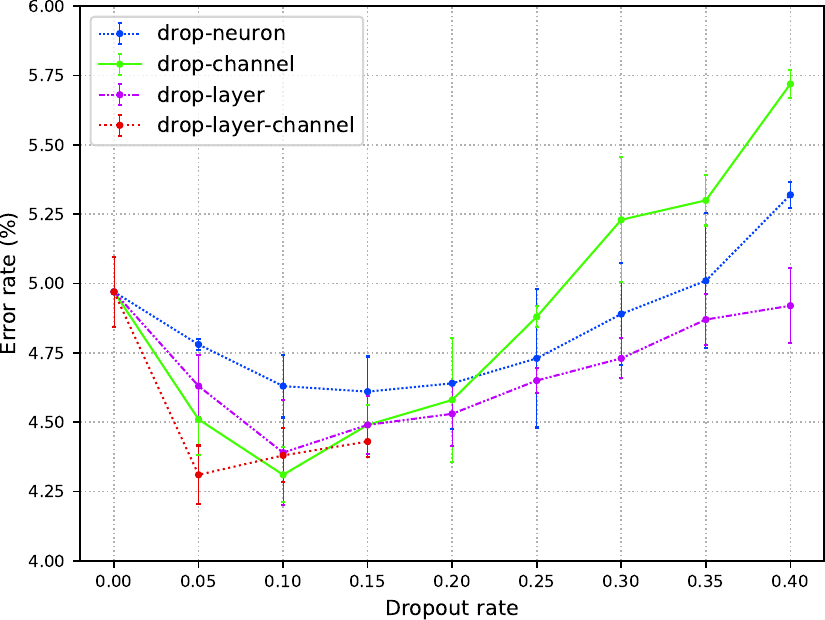}
    \caption{Error rate (\%) of WRN-40-4 trained with \dropneuron, \dropchannel and \droplayer.}
    \label{fig:wrn_10_all}
\end{figure}

To understand the impact of dropouts on the training process, we plot the learning curves trained with different dropout variants in Figure~\ref{fig:resnext_curve_all}.
We can notice that firstly, CNNs trained with different dropout operations achieves noticeably better results than the network without dropout.
Interestingly, the learning curve of networks with \droppath fluctuates drastically, though it achieves a lower error rate than other dropout variants.
However, when trained with \dropchannel, the training is more stable, and such integration yields the overall best result of a 4.49\% test error rate.
We conjecture that this is mainly because \droppath is a more radical regularization method, where each entire path is randomly dropped, and thus has higher variance.
Therefore, \droppath may need a lower dropout rate.

\begin{table*}[ht]
\centering
\begin{tabular}{ l | cc | cc | cc | c}
\thickhline
\multicolumn{1}{c|}{Model} & Depth & Params & C10 & C10+ & C100 & C100+ &SVHN \\
\hline

VGG~\cite{simonyan2014very} & 11 & 9.89M & 8.24 & 5.09 & 23.58 & 32.08 & - \\
\quad \myarrow with drop-neuron & 11 & 9.89M & 4.88 & 6.85 & 23.15 & 27.71 & - \\
\quad \myarrow with drop-channel & 11 & 9.89M & 4.62 & 7.76 & 21.89 & 29.51 & - \\
\hline
Wide ResNet~\cite{zagoruyko2016wide} & 40 & 8.95M & - & 4.97 & - & - & - \\
\quad \myarrow with drop-neuron & 40 & 8.95M & - & 4.61 & - & - & - \\
\quad \myarrow with drop-channel & 40 & 8.95M & - & 4.31 & - & - & - \\
\quad \myarrow with drop-layer & 40 & 8.95M & - & 4.39 & - & - & - \\
Wide ResNet~\cite{zagoruyko2016wide} & 16 & 10.96M & - & - & - & - & 1.54 \\
\quad \myarrow with drop-neuron & 16 & 10.96M & - & - & - & - & {\color{blue}\textbf{1.44}} \\
\hline

ResNeXt~\cite{xie2017aggregated} & 29 & 8.85M & - & 5.10 & - & - & - \\
\quad \myarrow with drop-neuron & 29 & 8.85M & - & 4.87 & - & - & - \\
\quad \myarrow with drop-channel & 29 & 8.85M & - & 4.72 & - & - & - \\
\quad \myarrow with drop-path & 29 & 8.85M & - & 4.62 & - & - & - \\
\hline

DenseNet-BC (k=12)~\cite{huang2016densely} & 100 & 0.8M & 5.92 & 4.51 & 24.15 & 22.27 & 1.76 \\
\quad \myarrow with drop-channel & 100 & 0.8M & 5.59 & 4.24 & 23.73 & 20.75 & 1.65 \\
DenseNet-BC (k=40)~\cite{huang2016densely}& 190 & 25.6M & - & 3.46 & - & 17.18 & -\\
\quad \myarrow with drop-neuron & 190 & 25.6M & - & 3.42 & - & 16.69 & -\\
\quad \myarrow with drop-channel & 190 & 25.6M & - & {\color{blue}\textbf{3.17}} & - & {\color{blue}\textbf{16.15}} & -\\

\thickhline

\end{tabular}
\caption{Overall results reported in error rate (\%) on CIFAR and SVHN datasets. A suffix + indicates standard data augmentation. Only results in Section~\ref{sec:experiment} are provided for comparison. The overall best results are highlighted in {\color{blue} \textbf{blue}}.} \label{tab:all_result_table}
\end{table*}

In particular, we further evaluate the effect of \droplayer with ResNet with Stochastic Depth~\cite{huang2016deep}, which is discussed in Section~\ref{sec:higher_level_dropout} and Section~\ref{general_mechanisms}.
We focus on comparing \droplayer with other dropout variants, namely \dropneuron and \dropchannel.
We adopt WRN-40-4 (see Table~\ref{tab:cnns_cifar}) with dropout rate from 0.0 to 0.40 in every 0.05.
The main results are summarized in Figure~\ref{fig:wrn_10_all}.
We find that dropout operations help obtain noticeably better results.
Specifically, \dropchannel achieves the best result of a 0.66\% reduction of test error rate from 4.97\% to 4.31\%.
When trained with \dropchannel, \droplayer improves performance considerably, which is comparable to \dropchannel alone.
We therefore conjecture that \dropchannel is a better dropout training choice, which achieves generally better results with more stable training.

\subsubsection{Dropout: Improvement over State-of-the-art CNNs}


\begin{table}[t!]
\centering
\begin{tabular}{ l | cc | c}
\thickhline
\multicolumn{1}{c|}{Model} & Depth & Params & ImageNet \\
\thickhline

VGG-16~\cite{simonyan2014very} & 16 & 138.4M & 27.63 \\
\quad \myarrow +drop-channel & 16 & 138.4M & 27.49 \\
\hline
WRN-50-2~\cite{zagoruyko2016wide} & 50 & 68.9M & 21.91 \\
\quad \myarrow +drop-layer+channel & 50 & 68.9M & 21.68 \\
\hline
DenseNet-L169-K32~\cite{huang2016densely} & 169 & 14.1M & 23.62 \\
\quad \myarrow +drop-path+channel & 169 & 14.1M & 23.47 \\
\thickhline

\end{tabular}
\caption{Comparison of Top-1 (single model and single crop) error rates on ImageNet classification dataset.} \label{tab:imagenet_table}
\end{table}

Thus far, we have examined the building blocks of different structural levels of dropout for regularizing the training of deep CNNs, and meanwhile evaluate the effectiveness of different dropout operations, i.e., \dropneuron, \dropchannel, \droppath (and \droplayer) with extensive experiments.
With these general and effective dropout training mechanisms, we can further improve the training of best-performing CNNs on benchmark datasets.
The overall results on CIFARs and SVHN are summarized in Table~\ref{tab:all_result_table}.

\begin{figure}[t!]
    \centering
    \includegraphics[width=0.42\textwidth]{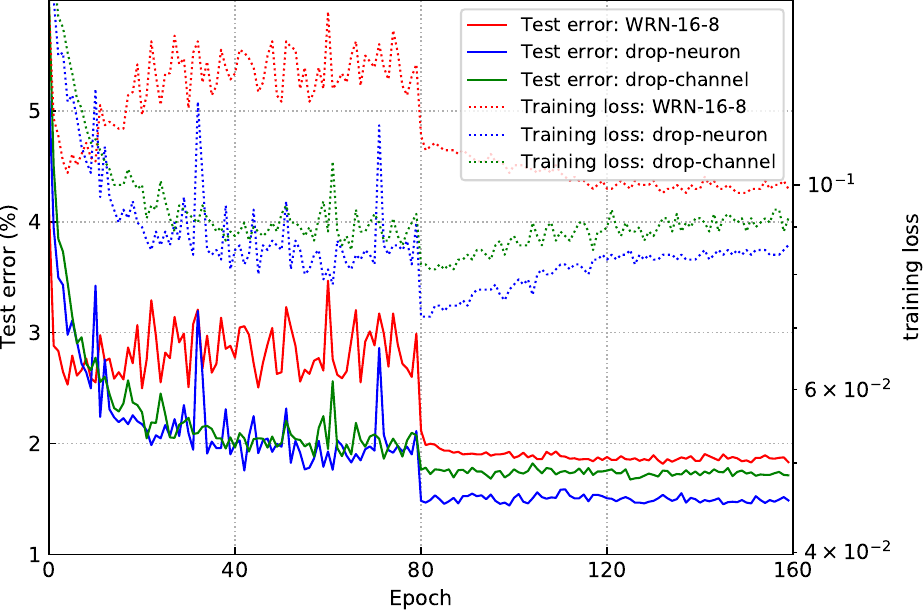}
    \caption{Learning curves and losses of WRN-16-8 trained with \dropneuron, \dropchannel on SVHN.}
    \label{fig:wrn_svhn_curve_full}
\end{figure}

For SVHN, WRN-16-8 (Table~\ref{tab:cnns_cifar}) originally achieves 1.54\% error rate without data augmentation.
We then introduce our proposed \dropneuron and \dropchannel building blocks to WRN, and the results are summarized in 
Figure~\ref{fig:wrn_svhn_curve_full}.
The results 
confirm that \dropneuron is more effective in regularizing networks trained without data augmentation.
Replacing conventional convolutional layers with the \dropneuron convolutional building blocks in WRN-16-8, we achieve a noticeably lower error rate of 1.44\% on SVHN over the original state-of-the-art model.
Figure~\ref{fig:wrn_svhn_curve_full} further shows that the proposed \dropneuron can effectively regularize the training process with a significantly lower training loss and meanwhile a lower error rate.
Without dropout regularization, the training stagnates quickly, and the test error rate even increases and fluctuates dramatically before the first drop of the learning rate.
While with dropouts, the training is more stable and the network continues to improve for lower errors.

\begin{figure*}[!ht]
    \centering
    \subfloat[CIFAR-100 w/o drop-neuron]{{\includegraphics[width=0.48\textwidth]{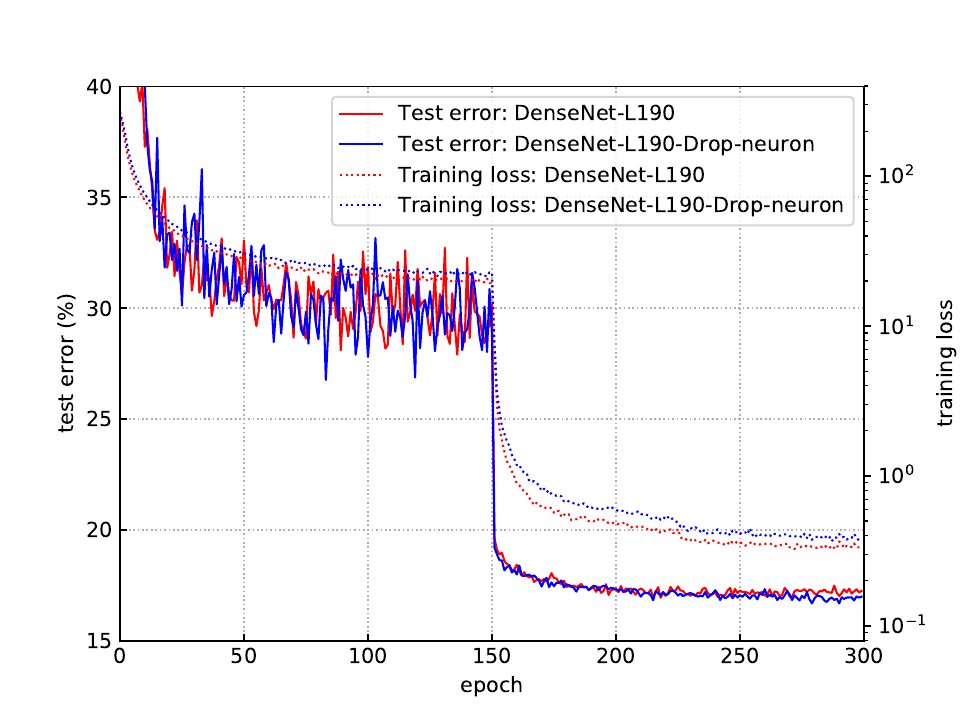} }}
    \subfloat[CIFAR-100 w/o drop-channel]{{\includegraphics[width=0.48\textwidth]{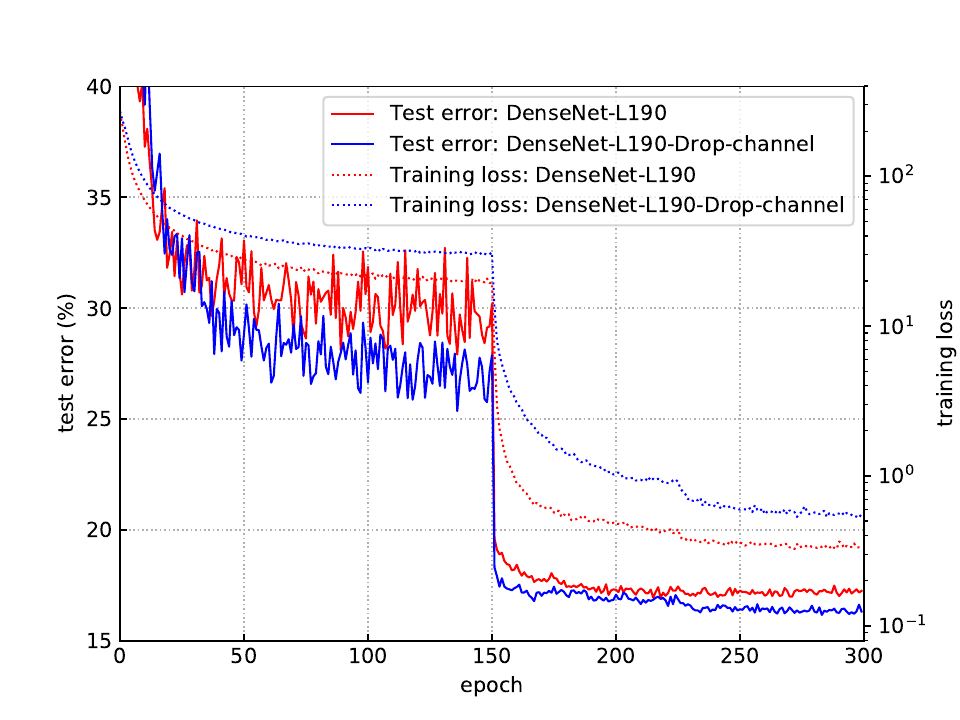} }}
    \caption{Test error and training loss curves for DenseNet-L190-K40 with and without \dropneuron and \dropchannel.
    The two corresponding dropout rates are both set to 0.1, leading to 16.76\% and 16.15\% test error rate respectively on CIFAR-100+.
    The test error is 17.18\% for the network trained without any dropout.}
    \label{fig:cifar_error_loss}
\end{figure*}

For CIFAR datasets, the state-of-the-art model DenseNet-L190-K40 (see Table~\ref{tab:cnns_cifar}) achieves 4.36\% and 17.18\% error rate on CIFAR-10 and CIFAR-100 respectively.
We then apply \dropneuron and \dropchannel building blocks in replacement of convolutional layers in the network, and with a dropout rate 0.1, DenseNet-L190-K40 achieves significantly better results of 3.17\% and 16.15\% error rate, i.e., a 0.29\% and 1.03\% relative error rate reduction respectively.

The overall experimental results on ImageNet dataset are summarized in Table~\ref{tab:imagenet_table}.
Three representative CNN architectures are adopted for the large dataset, specifically VGG-16~\cite{simonyan2014very} with the plain convolutional operation, WRN-50-2~\cite{zagoruyko2016wide} with residual connection and DenseNet-L169-K32~\cite{huang2016densely} with a dense connection between layers.
We evaluate WRN-50-2 with both \droplayer and \dropchannel, and meanwhile DenseNet-L169-K32 with both \droppath and \dropchannel.
For VGG-16, we improve accuracy by 0.14\% with \dropchannel.
For WRN-50-2, we observe a more significant error rate reduction of 0.23\% with the adoption of both \dropchannel and \droplayer.
With both \dropchannel and \droppath training, DenseNet-L169-K32 obtains a 0.15\% test error rate reduction.
The results of the three architectures on ImageNet further confirm that the dropout training mechanisms can significantly improve the training of deep CNNs if adopted properly.

Finally, to illustrate the difference between the regularization effect of \dropneuron and \dropchannel, we further plot training curves of the 190 layer DenseNet on CIFAR-100+ with the two dropout variants in Figure~\ref{fig:cifar_error_loss}.
The left panel indicates that the regularization effect of the \dropneuron training is rather limited, which is mainly because the channel instead of the neuron is the more suited structural level for the regularization of the convolutional transformation.
Compared with \dropneuron, \dropchannel regularizes the model effectively and thus achieves significantly better performance.
With \dropchannel, the test error decreases faster, and the training is more stable, especially before the first learning rate drop at epoch 150.
Furthermore, DenseNet regularized by \dropchannel learns with higher training loss while with much lower test error, indicating that \dropchannel prevents overfitting effectively.

\balance
\section{Conclusion}
\label{sec:conclusion}

In this paper, we formulated and examined the three structural levels of dropout training with a unified convolutional transformation framework, including \dropneuron, \dropchannel \droppath.
We attribute the failure of standard dropout to the incorrect placement in the convolutional building block, which incurs great training instability.
Through detailed discussion and analysis, we propose general convolutional building blocks supporting different structural levels of dropouts and a more effective variant \dropout, which are better in line with the convolutional transformation for CNNs and incurs negligible additional training costs.

Extensive analysis and experiments have shown that firstly, all these dropout variants are effective in improving the performance of convolutional neural networks by a noticeable margin.
Further, among these dropout methods, \dropneuron and \dropchannel are widely applicable to existing CNNs, while \droppath and particularly \droplayer are highly dependent on the CNN architecture.
In terms of effectiveness, \dropchannel generally outperforms other dropout variants.
This is largely due to the characteristic of the convolution transformation of CNNs, where the channel instead of other structural components is the most fundamental units participating in the operation.
Therefore, \dropchannel can better harness the benefits of both regularization and model ensemble.
We further note that \dropchannel could be more effective with Group Normalization.
Further, \dropneuron and \dropchannel can be integrated with higher levels of dropout variants, e.g., \droppath, which can further stabilize the training process.
On the other hand, \dropneuron outperforms \dropchannel in the network trained without data augmentation.

With the proposed building blocks designed for dropout training mechanisms, we can achieve noticeable improvement over state-of-the-art CNNs on CIFAR-10/100, SVHN and ImageNet datasets.
Given the generality and flexibility, these dropout training mechanisms would be useful for improving the training of a wide range of deep CNNs.

\IEEEpeerreviewmaketitle

\balance
\newpage

\bibliographystyle{my_abbrv}
\bibliography{reference}

\end{document}